\documentclass[journal]{IEEEtran}
\usepackage{amsmath,amsfonts}
\usepackage{algorithmic}
\usepackage{algorithm}
\usepackage{array}
\usepackage{amssymb}
\usepackage[caption,font=normalsize,labelfont=sf,textfont=sf]{subfig}
\usepackage{textcomp}
\usepackage{stfloats}
\usepackage{multirow}
\usepackage{url}
\usepackage{verbatim}
\usepackage{graphicx}
\graphicspath{ {./pic/} }
\usepackage{amsfonts}
\usepackage{bm}
\usepackage{cite}
\usepackage{booktabs, ragged2e}
\usepackage[flushleft]{threeparttable}
\usepackage{xcolor}

\newcommand\wu[1]{\textcolor{black}{#1}}

\begin{document}

\title{HighAir: A Hierarchical Graph Neural Network-Based Air Quality Forecasting Method}

\author{Ling Chen, Jiahui Xu, Binqing Wu, Mingqi Lv, Chaoqun Zhan, Sanjian Chen, and Jian Chang

\thanks{This work was supported in part by the National Key Research and Development Program of China under Grant 2018YFB0505000, in part by the Fundamental Research Funds for the Central Universities under Grant 2020QNA5017, and in part by Alibaba-Zhejiang University Joint Research Institute of Frontier Technologies. \textit{(Corresponding author: Ling Chen.)}}

\thanks{Ling Chen, Jiahui Xu, and Binqing Wu are with the College of Computer Science and Technology, Zhejiang University, 38 Zheda Road, Hangzhou 310027, China (emails: xujiahui19@zju.edu.cn, lingchen@cs.zju.edu.cn, binqingwu@cs.zju.edu.cn).}

\thanks{Mingqi Lv is with the College of Computer Science, Zhejiang University of Technology, 288 Liuhe Road, Hangzhou 310023, China (email: mingqilv@zjut.edu.cn).}

\thanks{Chaoqun Zhan, Sanjian Chen, and Jian Chang are with Alibaba Group, Hangzhou 311121, China (emails: lizhe.zcq@alibaba-inc.com, sanjian.c@alibaba-inc.com, jian.chang@alibaba-inc.com).}

}

\markboth{Journal of \LaTeX\ Class Files,~Vol.~14, No.~8, August~2015}%
{Shell \MakeLowercase{\textit{et al.}}: Bare Demo of IEEEtran.cls for IEEE Journals}

\maketitle

\begin{abstract}
Accurately forecasting air quality is critical to protecting general public from lung and heart diseases. This is a challenging task due to the complicated interactions among distinct pollution sources and various other influencing factors. Existing air quality forecasting methods cannot effectively model the diffusion processes of air pollutants between cities and monitoring stations, which may suddenly deteriorate the air quality of a region. In this paper, we propose HighAir, i.e., a hierarchical graph neural network-based air quality forecasting method, which adopts an encoder-decoder architecture and considers complex air quality influencing factors, e.g., weather and land usage. Specifically, we construct a city-level graph and station-level graphs from a hierarchical perspective, which can consider city-level and station-level patterns, respectively. We design two strategies, i.e., upper delivery and lower updating, to implement the inter-level interactions, and adopt message passing mechanism to implement the intra-level interactions. We dynamically adjust edge weights based on wind direction to model the correlations between dynamic factors and air quality. We compare HighAir with the state-of-the-art air quality forecasting methods on the dataset of Yangtze River Delta city group, which covers 10 major cities within 61,500 $\text{km}^2$. The experimental results show that HighAir significantly outperforms other methods.
\end{abstract}


\IEEEpeerreviewmaketitle

\section{Introduction}
\IEEEPARstart{A}{ir} quality is closely related to public health. As many cities have built monitoring stations, air quality data become increasingly available, and thus air quality forecasting has become a research hotspot for its potential benefit in supporting regulatory decision-making and public activity-planning.

Air quality forecasting involves multiple challenges. First, since the air pollution path of urban atmosphere consists of emission and diffusion processes \cite{rohde2015air, wang2016contribution, zhang2012real}, air quality has complex dependencies in spatial and temporal dimensions, which are difficult to capture. Second, air quality is affected by multi-source complex factors, e.g., weather and land usage \cite{zhang2018spatiotemporal}, and the corresponding knowledge needs to be extracted from data sources.

Existing methods for air quality forecasting can be roughly divided into two categories: physical model based methods and machine learning based methods. Physical model based methods exploit domain knowledge to simulate the physical and chemical processes of air pollutants, e.g., street canyon models \cite{rakowska2014impact, xie2005impact} and Gaussian plume models \cite{arystanbekova2004application}. These methods are usually based on domain knowledge and the generalization abilities of them are limited. Machine learning based methods leverage a data-driven process to learn complex relationships between inputs and outputs. To capture temporal dependencies, time-series analysis models were applied to air quality forecasting, e.g., auto-regressive based models \cite{chelani2006air, lee2012seasonal} and recurrent neural networks (RNNs) \cite{chen2019deep, du2019deep, liang2018geoman, qi2019hybrid}. To capture spatial dependencies, some methods use convolutional neural networks (CNNs) \cite{huang2018deep, lee2019hybrid} or graph neural networks (GNNs) \cite{qi2019hybrid, lin2018exploiting}. However, these methods cannot make full use of multi-city data. Most of these methods focus on citywide air quality forecasting, ignoring inter-city interactions. In real-world environment, the air pollutants diffused from adjacent cities may cause the deterioration of air quality in the target city. On the other hand, although a few number of these methods have considered multi-city data, they model the stations of all cities in a flat structure, ignoring the differences in the interactions between cities and between stations.

Hierarchical graph neural networks (HGNNs) \wu{\cite{mrowca2018flexible, nassar2018hierarchical, ying2018hierarchical, guo2021hierarchical, ma2022histgnn, ma2022hierarchical}} are a kind of GNNs that model complex systems from a hierarchical perspective, which capture the spatial dependencies with different granularities. However, applying existing HGNNs to forecast air quality is non-trivial, as they usually focus on the interactions for individual time slots and the unstable hierarchical structure they adopt cannot explicitly model the interactions between cities.

In this paper, we propose HighAir, a hierarchical graph neural network-based air quality forecasting method to forecast air quality. The key idea of HighAir is motivated by the following observations: (1) Monitoring stations and pollution sources, e.g., factories and automobiles, are typically distributed in clusters centered around the cities. (2) The patterns of air pollutant diffusion are different between cities/stations. (3) The diffusion processes of air pollutants are greatly impacted by dynamic factors, especially wind direction.

The main contributions of this work are summarized as follows:

(1) We propose HighAir, which constructs a city-level graph and station-level graphs from a hierarchical perspective and leverages a customized form of HGNN to forecast air quality. \wu{To the best of our knowledge, HighAir is the first work to do message passing within and between hierarchical graphs jointly considering dynamic factors.}

(2) We designed two strategies called upper delivery (UD) and lower updating (LU), which introduce the historical air quality information of adjacent cities into the station-level graphs to support effective message passing between stations in different cities.

(3) We design an edge weight updating method, which dynamically adjusts the weights of edges based on wind direction to model the correlations between dynamic factors and air quality.

(4) We evaluate HighAir on the dataset of Yangtze River Delta city group, and compare it with the state-of-the-art air quality forecasting methods. The experimental results show that HighAir significantly outperforms existing methods.

The rest of this paper is organized as follows. Section II introduces the related work. Section III gives the details of the proposed method. Section IV gives the details of the dataset, experiment settings, and experiment results. Conclusions and future work are presented in Section V.

\section{Related Work}
\subsection{Air Quality Forecasting}
Existing air quality forecasting methods can be roughly divided into two categories: physical model based methods and machine learning based methods. Physical model based methods simulate the physical and chemical processes of air pollutant emission and diffusion. For example, Rakowska et al. \cite{rakowska2014impact} introduced the street canyon model to simulate the diffusion processes of air pollutants in near-surface space. Arystanbekova et al.\cite{arystanbekova2004application} introduced the Gaussian plume model to forecast local pollution levels. However, these methods are usually based on domain knowledge and the generalization abilities of them are limited.

Machine learning based methods use a data-driven process to capture the dependencies in spatial and temporal dimensions and model the correlations between multi-source complex factors and air quality. Chelani et al. \cite{chelani2006air} introduced a time-series analysis based air quality forecasting method, which can consider the temporal dependencies between historical observations. Time-series analysis models make several key assumptions, i.e., the mean and variance of series are stationary over time, which are usually not true in air quality forecasting. Some researchers utilized general machine learning models, e.g., Support Vector Regression (SVR) \cite{nieto2013svm}, Random Forest \cite{yu2016raq}, XGBoost \cite{zamani2019pm2, pan2018application}, and tensor decomposition \cite{xu2019fine}, to model the correlations between multi-source factors and air quality. These methods do not customize the general models according to the specificities of air quality forecasting, so the forecasting accuracy greatly depends on data preprocessing and feature selection.

Deep learning can perform automatic feature learning, which avoids manual feature selection that relies on empirical assumptions. Du et al. \cite{du2019deep} proposed a method that uses one-dimensional CNNs to capture local temporal trends and leverages bi-directional LSTMs to extract long-term temporal features. Xu et al. \cite{xu2018spatio} introduced echo state networks to predict daily AQI series. Liang et al. \cite{liang2018geoman} proposed GeoMAN, a LSTM-based encoder-decoder method, which uses attention mechanism to capture the relationships between the pollutants observed by the same sensor and the same type of pollutants observed by different sensors. Yi et al. \cite{yi2018deep} proposed DeepAir that constructs subnets for multi-source data, whose outputs are integrated to forecast air quality. Considering the sparseness of stations in spatial distribution, Chen et al. \cite{chen2019deep} proposed a multi-task learning framework PANDA, which utilizes the air quality estimation task to assist the forecasting task.

Recently, due to the ability of modeling non-Euclidean distributed entities, GNNs have been applied in spatial-temporal forecasting. Qi et al. \cite{qi2019hybrid} proposed GC-LSTM, which constructs a graph based on the locations of stations and introduces graph convolution operations to model the relevance of pair-wise stations. Lin et al. \cite{lin2018exploiting} proposed GC-DCRNN, which combines the processes of graph convolution and sequence modeling by replacing the matrix multiplication in GRU with diffusion convolution operation. Xu et al. \cite{xu2020spatiotemporal} proposed ST-MFGCN, which constructs the graph-structured traffic road network to capture the vehicle emission spatiotemporal variation patterns. Geng et al. \cite{geng2019spatiotemporal} proposed STMGCN, which constructs multiple graphs and utilizes the global contextual information in temporal dimension to capture the spatial-temporal correlations. Guo et al. \cite{guo2019attention} proposed ASTGCN, which introduces attention mechanism in both spatial and temporal dimensions.

\subsection{Hierarchical Graph Neural Networks}

\begin{table}[]
\centering
\caption{The characteristics of HighAir and other methods}
\label{tab:table_1}
\begin{tabular}{cccc}
\toprule
Method & Structure & $\begin{gathered}\text { Intra-level }       \\
\text { Modeling }\end{gathered}$ & $\begin{gathered}\text { Inter-level }      \\
\text { Modeling }\end{gathered}$ \\
\midrule 
GC-LSTM     & flat GNN          & GCN                   & /                      \\
GC-DCRNN    & flat GNN          & diffusion conv.       & /                      \\
HRN         & stable HGNN       & FNN                   & FNN                   \\
ST-UNet     & unstable HGNN     & GCN                   & max/mean pooling      \\
\wu{HGCN}     & unstable HGNN     & GCN                   &  \wu{fix assignment}     \\
\wu{HiSTGNN}     & unstable HGNN     & GCN                   & \wu{fix assignment}      \\
\wu{HiGNN}     & unstable HGNN     & GCN                   & \wu{learnable assignment}      \\
HighAir     & stable HGNN       & message passing       & UD/LU                 \\
\bottomrule
\end{tabular}
\end{table}

HGNNs are a kind of GNNs that model complex systems from a hierarchical perspective, which captures the spatial dependencies with different granularities. The principle of HGNNs is to generate the coarsened graph from an original graph and implement interactions between them. Based on the strategy of generating the coarsened graph, HGNNs can be roughly divided into two categories: stable HGNN and unstable HGNN.

In the first category, super nodes (i.e., the nodes in the coarsened graph) and the connections between them are fixed or relatively stable. Li et al. \cite{li2019semi} proposed SEAL-C/AI, which divides social networks into  user group level and individual user level, and constructs two classifiers with different views to solve the node classification problem. Mrowca et al. \cite{mrowca2018flexible} proposed HRN, which decomposes a physical object into multi-level entities to simulate the physical dynamics of the object.

In the second category, super nodes and the connections between them evolve with the node attributes of the original graph. The coarsened graph is generated based on manually designed rules \cite{hu2019hierarchical, yu2019st}, node clustering \cite{nassar2018hierarchical}, \wu{and spectral clustering \cite{guo2021hierarchical}}. Yu et al. \cite{yu2019st} proposed ST-UNet, which uses paired sampling operation to get multi-resolution graphs and adopts dilated RNNs to capture temporal dependencies. \wu{Ma et al. \cite{ma2022histgnn} proposed HiSTGNN, which constructs two-layer graphs by adopting factor and node embeddings (parameters of networks), one for weather factors and the other for stations. Ma et al. \cite{ma2022hierarchical} proposed HiGNN, which constructs dynamic county-level and region-level mobility graph structures via mobility flows and a learnable assignment matrix.}

Table \ref{tab:table_1} summarizes the characteristics of existing methods and HighAir. HighAir is a stable hierarchical structured method, which can explicitly model the interactions between cities. Different from some HGNN methods focusing on the interactions for individual time slots, the UD and LU strategies of HighAir introduce the historical air quality information of adjacent cities into the station-level graphs to support effective message passing between stations in different cities.

\section{Methodology}

\subsection{Definitions}
\textit{Definition I (Cities and monitoring stations)}: We define $C=$ $\left\{c_a, 1 \leq a \leq N\right\}$ as the set of $N$ cities, $L=\left\{\boldsymbol{l}_a, 1 \leq a \leq N\right\}$ as the set of the locations of these cities, $S_a=\left\{s_{a, i}, 1 \leq i \leq\left|S_a\right|\right\}$ as the set of monitoring stations in city $c_a$, and $\mathcal{L}_a=\left\{\boldsymbol{l}_{a, i}, 1 \leq\right.$ $\left.i \leq\left|S_a\right|\right\}$ as the set of the locations of monitoring stations in city $c_a$. $\boldsymbol{l}_{a, i}$ consists of the latitude and longitude of a station. $\boldsymbol{l}_a$ is defined as the average value of $\boldsymbol{l}_{a, i}$ in $\mathcal{L}_a$.

\begin{figure}[]
    \centering
    \captionsetup{justification=centering}
    \includegraphics[width = 0.25\textwidth]{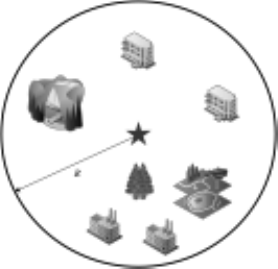}
    \caption{The POI distribution of a station.}
    \label{fig_1}
\end{figure}

\textit{Definition 2 (POI data)}: Points of Interest (POI) data contain the land usage information of a region, which consider five POI categories: residential, park, mountain, water (river or pool), and industrial. As shown in Fig. \ref{fig_1}, the star represents a monitoring station $s_{a, i}$. There are two residential areas, one park, one mountain, one pool, and two industrial facilities within the perception radius $R$ of $s_{a, i}$. Thus, the POI vector $\boldsymbol{poi}_{a, i}$ can be represented as $[2,1,1,1,2]$.

\textit{Definition 3 (Weather data)}: The weather data of city $c_a$ at time slot $t$ is represented as a vector $\boldsymbol{weather}_a^t$, including temperature, humidity, rainfall, wind speed, wind direction, and air pressure. We encode the wind direction data as a vector $\boldsymbol{wind}_a^t$ according to the encoding rules shown in Table \ref{tab:table_2}. Especially, the weather data of future time slots are obtained from weather forecast.

\begin{table}[]
\centering
\caption{The encoding rules of wind direction}
\label{tab:table_2}
\resizebox{.25\textwidth}{!}{
\begin{tabular}{ll}
\toprule
Direction              & Vector      \\ 
\midrule
North                  & {[}0,1{]}   \\
Northeast              & {[}1,1{]}   \\
East                   & {[}1,0{]}   \\
Southeast              & {[}1,-1{]}  \\
South                  & {[}0,-1{]}  \\
Southwest              & {[}-1,-1{]} \\
West                   & {[}-1,0{]}  \\
Northwest              & {[}-1,1{]}  \\
No sustained direction & {[}0,0{]}   \\ 
\bottomrule
\end{tabular}}
\end{table}

\textit{Definition 4 (AQI data)}: The Air Quality Index (AQI) of station $s_{a, i}$ at time period $t$ is represented as $\boldsymbol{aqi}_{a, i}^t$.

\textit{Definition 5 (Air quality forecasting task)}: Given city locations $L$, station locations $\mathcal{L}$, POI data $\boldsymbol{poi}$, $\tau_{\text{in}}$ hours of AQI data $\boldsymbol{aqi}$, $\tau_{\text {in }}+\tau_{\text {out }}$ hours of weather data $\boldsymbol{weather}$, the air quality forecasting task aims to forecast the AQIs of monitoring stations for the next $\tau_{\text {out }}$ hours, where $\tau_{\text {in }}$ denotes the length of historical time window and $\tau_{\text {out }}$ denotes forecasting horizon.

\begin{figure*}
    \centering
    \includegraphics{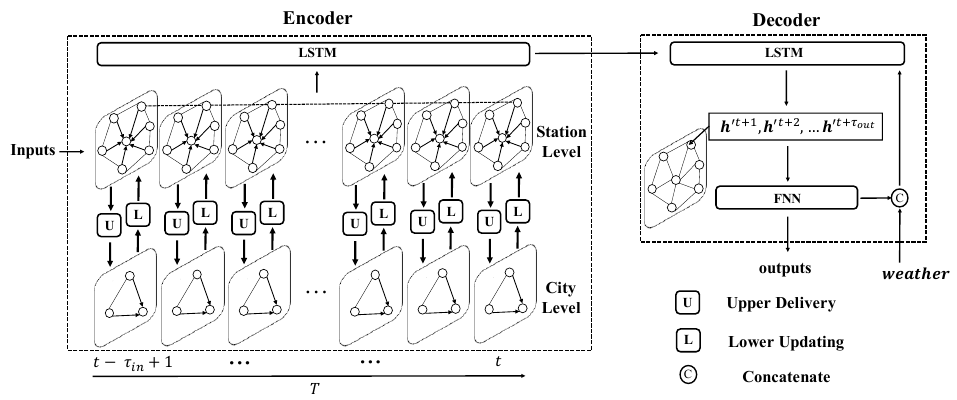}
    \caption{The framework of HighAir.}
    \label{fig:framework}
\end{figure*}

\subsection{Framework}
Fig. \ref{fig:framework} shows the framework of HighAir, which adopts an encoder-decoder architecture. For the encoder, HighAir constructs a city-level graph and station-level graphs from a hierarchical perspective. The model adopts message passing mechanism to implement the intra-level interactions between cities/stations. Upper delivery and lower updating are designed to implement the inter-level interactions. For each time slot, HighAir computes the node attributes in the city-level graph, which would be utilized to update the global attributes of the corresponding station-level graphs. After the message passing process on the station-level graphs, the node attribute sequence of a station would be sent to encoder LSTM. For the decoder, HighAir uses weather data $\boldsymbol{weather}$ to assist forecasting task. Each hidden state of decoder LSTM is treated as the input of a fully-connected neural network (FNN) that outputs the forecasting result of the corresponding time slot.

\subsection{Graph Construction}
Existing air quality forecasting methods construct a flat, static graph to model the interactions between stations \cite{qi2019hybrid, lin2018exploiting}. However, messages between stations located in different cities are hard to pass on a flat graph, as the distances between these stations are too larger as compared with the distances between adjacent stations. Moreover, dynamic factors, e.g., wind direction, cannot be effectively utilized in static graphs with fixed edge weights.

We use node set $V$ to represent cities/stations in the city-level graph/the station-level graphs, edge set $E$ to represent the relevance between nodes, and global attribute vector $\boldsymbol{u}$ to represent the global state of a station-level graph.

We introduce geographic similarity $gs$ and wind direction similarity $ws$ to measure static and dynamic relevance between nodes, respectively. Geographic similarity $gs$ is calculated using the geographic distance between cities/stations, which is symmetric. Wind direction similarity $ws$ is dynamically adjusted by the real-time wind directions and relative locations between cities/stations, which is asymmetric.

Given cities/stations $a$ and $b, \boldsymbol{l}_a=\left[x_a, y_a\right]$ and $\boldsymbol{l}_b=$ $\left[x_b, y_b\right]$ represent the locations of $a$ and $b$. The geographic similarity $g s$ of $a$ and $b$ is defined as follows:

\begin{equation}
d_{a, b}=d_{b, a}=\sqrt{\left(x_a-x_b\right)^2+\left(y_a-y_b\right)^2}\
\end{equation}

\begin{equation}
g s_{a, b}
=g s_{b, a}
=\left\{\begin{array}{c}
\frac{1}{d_{a, b}}, 0<d_{a, b}<R_h \\
0, {otherwise}
\end{array}\right.
\end{equation}
where $d_{a, b}$ and $d_{b, a}$ denote the Euclidean distance between $a$ and $b$, and $R_h$ is the geographic distance threshold. Considering the differences of different cities, we adopt an adaptive method to determine $R_h$ for each graph, which is defined as follows:

\begin{equation}
\mathcal{D}_a={min} \left(d_{a, k}\right), a, k \in V, a \neq k 
\end{equation}

\begin{equation}
D={max} \left(\mathcal{D}_a\right), a \in V
\end{equation}

\begin{equation}
R_h=\lambda D, \lambda \geq 1   
\end{equation}

where $\mathcal{D}_a$ denotes the minimum distance between node $a$ and other nodes, $D$ denotes the maximum $\mathcal{D}$ value of all nodes, and distance weight $\lambda$ is a hyper-parameter to control the number of edges in the constructed graph. In this way, all nodes in the constructed graph have at least one edge connected to other nodes, and $R_h$ should not be too large to make the graph fully-connected.

To capture the diffusion processes of pollutants between cities/stations according to wind directions, the wind direction similarity $w s$ between $a$ and $b$ at time period $t$ is defined as follows:

\begin{equation}
 \boldsymbol{\zeta_{a \rightarrow b}}
 =\boldsymbol{l}_b-\boldsymbol{l}_a, \boldsymbol{\zeta_{b \rightarrow a}}
 =\boldsymbol{l}_a-\boldsymbol{l}_b   
\end{equation}

\begin{equation}
 w s_{a, b}^t=
 \frac{\boldsymbol{wind}_a^t \cdot \boldsymbol{\zeta_{a \rightarrow b}}}
 {\left|\boldsymbol{wind}_a^t\right|\left|\boldsymbol{\zeta_{a \rightarrow b}}\right|}    
 \end{equation}
 
\begin{equation}
w s_{b, a}^t=
\frac{\boldsymbol{wind}_b^t \cdot \boldsymbol{\zeta_{b \rightarrow a}}}
{\left|\boldsymbol{wind}_b^t\right|\left|\boldsymbol{\zeta_{a \rightarrow b}}\right|}    
\end{equation}
where $\cdot$ denotes inner product operation, and $\boldsymbol{wind}_a^t$ and $\boldsymbol{wind}_b^t$ denote the wind direction vectors of $a$ and $b$ at time period $t$. According to the above definitions, the value range of $ws$ is $[-1,1]$. The larger $w s_{a, b}^t$ is, the more likely the air pollutants from city $a$ would be brought to city $b$ by wind at time period $t$.

\begin{figure}[]
    \centering
    \includegraphics{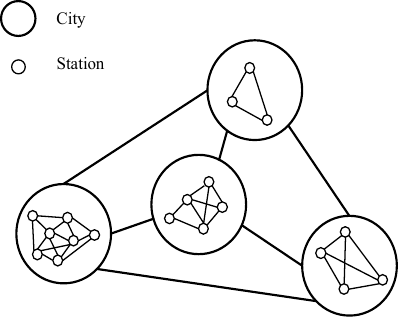}
    \caption{The city-level graph and the station-level graphs with a hierarchical structure.}
    \label{fig_3}
\end{figure}

At this point, we have constructed the city-level graph and the station-level graphs with a hierarchical structure, as shown in Fig. \ref{fig_3}. Compared with flat graphs, nodes in the city-level graph are graph instances, which can support effective message passing between stations in different cities.

\begin{figure*}[]
    \centering
    \includegraphics{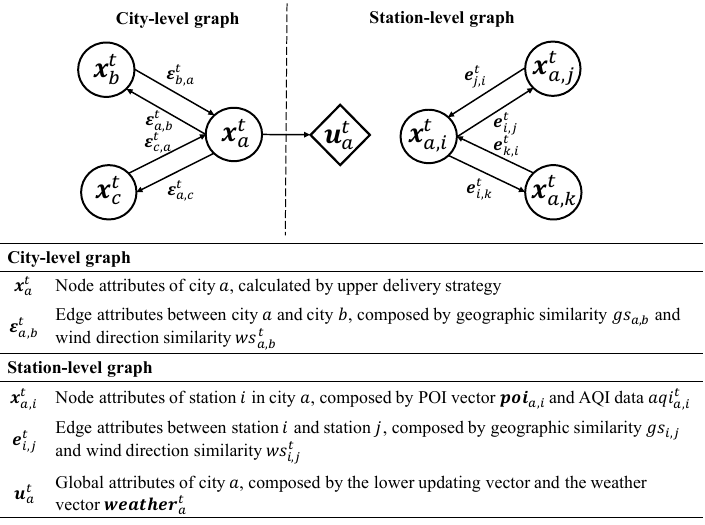}
    \caption{The attributes of the city-level graph and the station-level graphs.}
    \label{fig_4}
\end{figure*}

Fig. \ref{fig_4} gives the attributes of the city-level graph and the station-level graphs. Node attributes $\boldsymbol{x}^t$ denotes the monitoring data about a city (i.e., AQI data) or station (i.e., AQI and POI data) at time period $t$. Edge weights $\boldsymbol{\varepsilon}^t / \boldsymbol{e}^t$ denotes the similarity between two cities/stations at time period $t$. Global attributes $\boldsymbol{u}_a^t$ denotes the common features of stations in city $a$ at time period $t$. All the components of node attributes, edge weights, and global attributes are combined by concatenation.

\subsection{The Modeling of Spatial Dependencies}
HGNNs can model complex systems from a hierarchical perspective, which can capture spatial dependencies under different granularities. We adopt message passing mechanism to capture the intra-level interactions. In addition, we design two strategies, called upper delivery and lower updating, to model the messages passing from station-level to city-level and city-level to station-level, respectively. For convenience, the edge weights in the city-level graph and station-level graphs are all denoted by e in the rest of this paper.

\textit{1) The Modeling of Intra-Level Interactions}: To implement the intra-level interactions on the city-level graph and the station-level graphs, we adopt message passing mechanism \cite{battaglia2018relational}, which allows cities/stations to perceive information about the adjacent cities/stations. The mechanism consists of two major processes: message aggregation and representation updating, which are defined as follows (the superscript $t$ is omitted for neatness):

\begin{equation}
\boldsymbol{\gamma}_{a,s}=\sigma_a\left(\boldsymbol{W}_a\left(\boldsymbol{x}_s\left\|\boldsymbol{x}_a\right\| \boldsymbol{e}_{a, s}\right)+\boldsymbol{b}_a\right), s \in \mathcal{N}(a)    
\end{equation}

\begin{equation}
 \boldsymbol{r}_a={mean}\left(\boldsymbol{\gamma}_{a, s}\right), s \in \mathcal{N}(a)  
\end{equation}

\begin{equation}
 \begin{aligned}
 \left\{\begin{array}{c}
\boldsymbol{x}_a^{\prime}=\sigma_c\left(\boldsymbol{W}_c\left(\boldsymbol{r}_a \| \boldsymbol{x}_a\right)+\boldsymbol{b}_c\right), {city\ graph } \\
\boldsymbol{x}_{a, i}^{\prime}=\sigma_s\left(\boldsymbol{W}_s\left(\boldsymbol{r}_{a, i}\left\|\boldsymbol{x}_{a, i}\right\| \boldsymbol{u}_a\right)+\boldsymbol{b}_s\right), {station\  graphs}
\end{array}\right.     
\end{aligned}   
\end{equation}
where $\boldsymbol{x}_a$ is the attribute of node $a, \boldsymbol{x}_s$ is the attribute of a neighbor node $s, \boldsymbol{e}_{a, s}$ is the edge weight, $\boldsymbol{\gamma}_{a, s}$ denotes the messages passed from node $s$ to node $a, \boldsymbol{u}_a$ denotes the global attributes of city $a$, and $\boldsymbol{r}_a$ denotes the aggregated message passed from the neighbors of node $a . \boldsymbol{W}_a, \boldsymbol{W}_c, \boldsymbol{W}_s, \boldsymbol{b}_a, \boldsymbol{b}_c, \boldsymbol{b}_s$ are learnable parameters, $\sigma_a, \sigma_c, \sigma_s$ are activation functions, $||$ denotes the concatenation operation, and $mean$ denotes the mean aggregation method. For the station-level graphs, the parameters are not shared among different cities.

In this way, the model can capture the diffusion processes of pollutants between cities/stations based on geographic similarity gs and wind direction similarity $ws$. We construct a station-level graph for each city.

\textit{2) Upper Delivery Strategy}: Upper delivery is designed to model the interactions from the station-level graphs to the city-level graph. The messages passed from the station-level graphs contain the current and historical air quality information.

There are two optional methods to get the overall air quality representation of a city: mean operation, which takes the mean AQI value of all stations in the city as the representation, and SAGE proposed by Li et al. \cite{li2019semi}, which considers the graph structure and automatically learns the weights of different stations in the city. Considering that mean operation is a non-parametric method, we use it here (the corresponding comparison is given in Section IV-F), which is defined as follows:

\begin{equation}
 {AQI}_a^t={mean}\left({aqi}_{a, i}^t\right), 1 \leq i \leq\left|S_a\right|   
\end{equation}

For each city, we calculate the overall air quality representation of the city at each time slot, forming a sequence $\left\{A Q I_a^{t-\tau_{\mathrm{in}}+1}, A Q I_a^{t-\tau_{\mathrm{in}}+2}, \ldots, A Q I_a^t \mid 1 \leq a \leq N\right\}$, which would be sent to the city LSTM to get the representations of current and historical air quality information. The city LSTMs share parameters among different cities.

\begin{figure}[]
    \centering
    \includegraphics{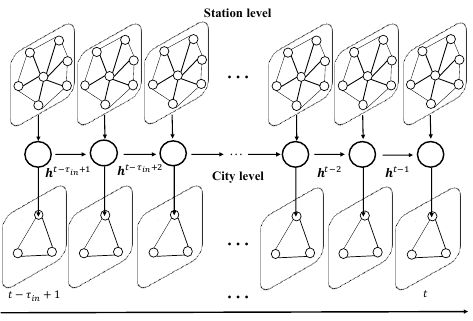}
    \caption{The upper delivery strategy.}
    \label{fig_5}
\end{figure}

We take the hidden state $\boldsymbol{h}$ of the city LSTM at each time slot as the initial attributes of the corresponding node in the city-level graph. Fig. \ref{fig_5} shows the process of upper delivery. In this way, the representation of a city contains both the current and historical air quality information, i.e., cities on the graph can perceive the historical air quality information of adjacent cities.

\textit{3) Lower Updating Strategy}: Lower updating is designed to model the interactions from the city-level graph to the station-level graphs. The messages passed from the city-level graph contain the historical air quality information of adjacent cities, which has been achieved by the message passing process on the city-level graph.

\begin{figure}[]
    \centering
    \includegraphics{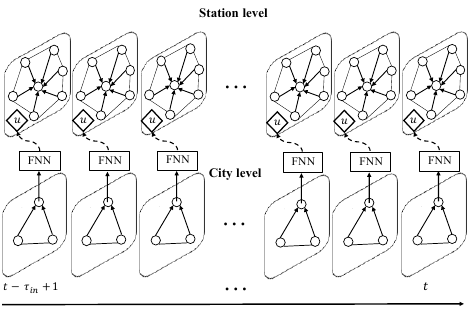}
    \caption{The lower updating strategy.}
    \label{fig_6}
\end{figure}

As shown in Fig. \ref{fig_6}, after the message passing process on the city-level graph at each time slot, the representation of a city at each time slot is fed to a FNN to get a lower updating vector, which would be used to update the global attributes $u$ of the corresponding station-level graph. The FNNs share parameters among different cities.

The global attributes of city $a$ station-level graph $\boldsymbol{u}_a^t$ consist of weather data $\boldsymbol{weather}_a^t$ and the lower updating vector at time period $t$, and global attributes would be used in the message passing process on the station-level graph (Section III-D1). In this way, stations can perceive the historical air quality information of adjacent cities during the message passing process on the station-level graphs.

\subsection{The Modeling of Temporal Dependencies}

\begin{figure}[]
    \centering
    \includegraphics{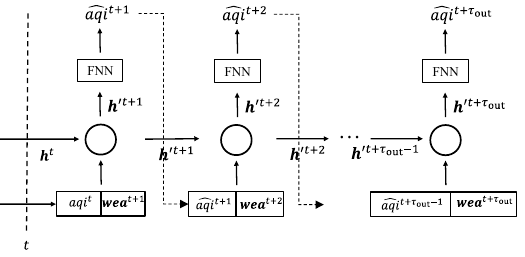}
    \caption{The decoder architecture of HighAir.}
    \label{fig_7}
\end{figure}

To capture temporal dependencies, HighAir adopts an encoder-decoder architecture. For each station, the corresponding node attributes on the station-level graphs form a sequence
$X_{a,i} = \{\boldsymbol{x}_{a,i}^{t-\tau_{\text{in}}+1}, \boldsymbol{x}_{a,i}^{t-\tau_{\text{in}}+2}, \dots, \boldsymbol{x}_{a,i}^{t} | 1 \leq a \leq N, 1 \leq i \leq |S_a|\}$ (each time slot in the historical time window has a corresponding node attribute). The encoder LSTM takes $X_{a,i}$ as input, and the last hidden state of the encoder LSTM is reserved to initialize the hidden state of the decoder LSTM. As shown in Fig. \ref{fig_7}, the input of the decoder LSTM consists of the last forecasted AQI and the weather data $\boldsymbol{weather}$ of the corresponding time slot. Each hidden state of the decoder LSTM would be taken as the input of a FNN, which outputs the forecasting result of the corresponding time slot.

The encoder LSTMs (as well as the decoder LSTMs and the FNNs) of different stations in a city share parameters. In this way, we can forecast the air quality for the next $\tau_{\text{out}}$ time slots ($\tau_{\text{out}}$ denotes forecasting horizon).

\subsection{Learning}
The air quality forecasting task aims to get the forecasted AQIs of the monitoring stations for the next $\tau_{\text {out}}$ time slots. We calculate the mean square error between the true AQIs and forecasted AQIs to measure the errors of forecasted results. We define the loss function as follow:

\begin{equation}
L(\theta)=\frac{1}{\tau_{\text {out }} \Sigma_{a=1}^{|C|}\left|S_a\right|} \sum_{a=1}^{|C|} \sum_{i=1}^{\left|S_a\right|} \sum_{k=1}^{\tau_{\text {out }}}\left|a q i_{a, i}^{t+k}-\widehat{a q \tau_{a, i}} t^{t+k}\right|^2    
\end{equation}
where $C$ denotes the set of cities whose air quality needs to be forecasted, $\left|S_a\right|$ denotes the station number of city $c_a, \widehat{aqi}_{a, i}^{t+k}$ denotes the forecasted AQI of station $s_{a, i}$ at time slot $t+k$, $aqi_{a, i}^{t+k}$ denotes the true AQI of station $s_{a, i}$ at time slot $t+k$, and $\theta$ denotes all the parameters used in HighAir.

In summary, the calculation process of HighAir is given by Algorithm 1. We first construct a city-level graph and station-level graphs with a hierarchical structure. We then apply upper delivery and lower updating strategies to introduce the historical air quality information of adjacent cities to station level, and adopt the message passing mechanism to update the representation of each station. To further model the temporal dependencies, we adopt LSTM-based encoder-decoder architecture to forecast the AQI data of each station.

\begin{table*}[h]
\centering
\caption{The station number of each city}
\label{tab:table_3}
\begin{tabular}{@{}lllllllllll@{}}
\toprule
City           & Hangzhou               & Shaoxing              & Ningbo                & Zhoushan              & Jiaxing               & Huzhou                & Shanghai               & Suzhou                & Nantong       & Wuxi                  \\ 
\midrule
Station Number & \multicolumn{1}{c}{10} & \multicolumn{1}{c}{2} & \multicolumn{1}{c}{7} & \multicolumn{1}{c}{3} & \multicolumn{1}{c}{2} & \multicolumn{1}{c}{1} & \multicolumn{1}{c}{10} & \multicolumn{1}{c}{8} & \multicolumn{1}{c}{5} & \multicolumn{1}{c}{8} \\ 
\bottomrule
\end{tabular}
\end{table*}

\begin{figure}[]
    \centering
    \includegraphics{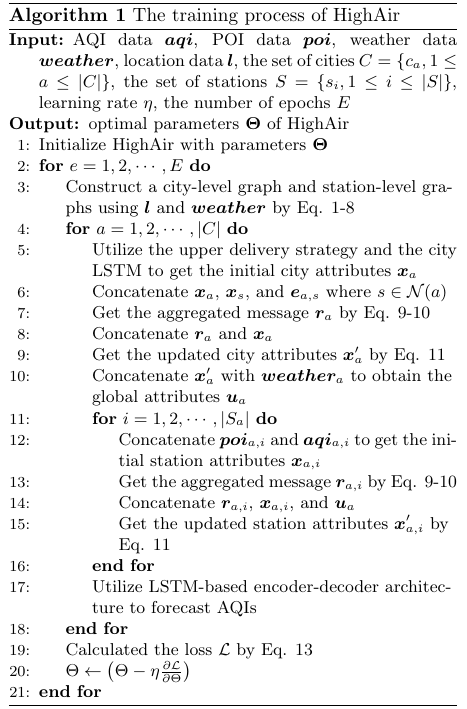}
    \label{fig_algorithm}
\end{figure}

\section{Experiments}
In this section, we present the experiments to evaluate HighAir. The detailed descriptions of the dataset and settings are provided in Section IV-A and Section IV-B, respectively. We investigate how distance weight $\lambda$, the number of message passing layers at city-level $L_{\text {city}}$, and the number of message passing layers at station-level $L_{\text {station}}$ affect the performance of HighAir in Section IV-C. The effectiveness of data used and method components would be discussed in Section IV-D and Section IV-E, respectively. We compare our method with other air quality forecasting methods in Section IV-F, and provide a case study to illustrate the superior performance of HighAir in Section IV-G.

We introduce two metrics: mean absolute error (MAE) and root mean squared error (RMSE), to evaluate the performances of methods, which are defined as follows:

\begin{equation}
M A E  =\frac{1}{\sum_{a=1}^{|C|}\left|S_a\right|} \sum_{a=1}^{|C|} \sum_{i=1}^{\left|S_a\right|} \mid a q i_{a, i}^{t+k}-\widehat{a q i}_{a, i}^{t+k} \mid
\end{equation}

\begin{equation}
 R M S E  =\sqrt{\frac{1}{\sum_{a=1}^{|C|}\left|S_a\right|} \sum_{a=1}^{|C|} \sum_{i=1}^{\left|S_a\right|}\left|a q i_{a, i}^{t+k}-\widehat{a q i}_{a, i}^{t+k}\right|^2}    
\end{equation}

In the experiments, we utilize previous 24-hour observations to forecast the next 12-hour AQIs, i.e., $\tau_{\text {in }}=24$ and $\tau_{\text {out }}=12$, and select the results of 1 hour, 3 hours, 6 hours, and 12 hours ahead forecasting to report. All experiments are repeated 5 times to avoid contingency.

\subsection{Experimental Dataset}
\begin{figure}[]
    \centering
    \includegraphics{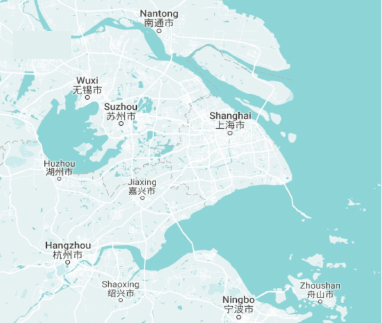}
    \caption{The Yangtze River Delta city group.}
    \label{fig_8}
\end{figure}

We evaluate the performance of HighAir on the dataset of Yangtze River Delta city group. Air quality situation in this region is very complicated, as the region is industrially developed and densely populated. The city group contains ten cities: Hangzhou, Ningbo, Shaoxing, Jiaxing, Zhoushan, Huzhou, Shanghai, Suzhou, Nantong, and Wuxi, as shown in Fig. \ref{fig_8}. The station number of each city is given in Table \ref{tab:table_3}.

The dataset contains air quality data, POI data, and weather data, which are collected from January 1 to December 21, 2018. The details of these data are described as follows:

(1) Air quality data, including the AQIs and the locations of each station, are collected from National Urban Air Quality Real-time Release Platform\footnote{http://106.37.208.233:20035/}. We collect AQI data at 1-hour granularity.

(2) POI data are collected from the map engine of AMAP\footnote{https://lbs.amap.com/api/webservice/guide/api/search/}. The perception radius R is set to 500 m.

(3) Weather data are collected from Envicloud\footnote{http://www.envicloud.cn/} and weather China\footnote{http://www.weather.com.cn/}, two data service providers. We collect weather data at 1-hour granularity.

We evaluate the performance of HighAir on the stations of Shanghai city, Suzhou city, and Jiaxing city. These cities are located in the center of the city group, and the information of adjacent cities is available. We use interpolation method to fill the missing values in the dataset. Sliding windows (step = 1 hour) are used to generate samples, and we finally get 8448 samples.

\subsection{Experimental Settings}
We split the generated samples into training data, validation data, and test data by the ratio of 0.7:0.1:0.2 in chronological order. We choose Adam \cite{kingma2014adam} as the optimizer in the training phase. Through hyper-parameter tuning (Section IV-C), distance weight $\lambda$, the number of message passing layers at city-level $L_{\text{city}}$, and the number of message passing layers at station-level $L_{\text {station}}$ are set to 1.2, 1, and 1. During the training phase, the batch size is set to 128, the epoch number is set to 300, the hidden size of GNNs is set to 32, the dimension of global attributes is set to 24, and the hidden state size of LSTMs is set to 64. Since all calculation processes are differentiable, HighAir can be trained via the back-propagation method. 

\begin{table*}[]
\centering
\caption{The evaluation results of $L_{\text {city}}$ and $L_{\text {station}}$ (mean$\pm$std)}
\label{tab:table_4}
\resizebox{.65\textwidth}{!}{
\begin{tabular}{lcccccc}
\toprule
$c$ & $L_{\text {station}}$ & Metric & 1h & 3h & 6h & 12h \\
\midrule

\multirow{2}{*}{1} & \multirow{2}{*}{1} 
& MAE       &\textbf{6.65}$\pm$\textbf{0.135}             &\textbf{13.15}$\pm$\textbf{0.382}      &\textbf{19.18}$\pm$\textbf{0.568}      &24.16$\pm$0.583 \\
& & RMSE    &\textbf{9.97}$\pm$\textbf{0.193}        & \textbf{19.01}$\pm$\textbf{0.516}      &\textbf{26.97}$\pm$\textbf{0.821}       &33.41$\pm$0.759 \\

\cline { 3 - 7 } 
\multirow{2}{*}{1} 
&\multirow{2}{*}{2} 
& MAE       &7.11$\pm$0.188       & 13.48$\pm$0.039        & 19.48$\pm$0.279       &24.62$\pm$0.388 \\
& & RMSE    &10.65$\pm$0.393    & 19.46$\pm$0.153         & 27.43$\pm$0.357         &33.89$\pm$0.386 \\

\cline { 3 - 7 } 
\multirow{2}{*}{1} 
&\multirow{2}{*}{3} 
& MAE           & 7.11$\pm$0.082      & 13.54$\pm$0.209         & 19.57$\pm$0.346         & 24.88$\pm$0.551 \\
& & RMSE        & 10.78$\pm$0.189   & 19.61$\pm$0.258           & 27.55$\pm$0.513         & 34.23$\pm$0.696 \\

\cline { 3 - 7 } 
\multirow{2}{*}{2} & \multirow{2}{*}{1}
& MAE           & 6.96$\pm$0.148            & 13.27$\pm$0.373       & 19.19$\pm$0.745       & 24.25$\pm$0.880 \\
& & RMSE        & 10.49$\pm$0.269           & 19.31$\pm$0.437       & 27.11$\pm$1.032       & 33.51$\pm$1.289 \\

\cline { 3 - 7 } 
\multirow{2}{*}{2} & \multirow{2}{*}{2}
& MAE           &6.90$\pm$0.115            &13.50$\pm$0.254           &19.61$\pm$0.400         &24.50$\pm$0.621 \\
& & RMSE        &10.26$\pm$0.155           &19.47$\pm$0.329           &27.77$\pm$0.640         &33.95$\pm$0.748 \\

\cline { 3 - 7 } 
\multirow{2}{*}{2} & \multirow{2}{*}{3}
& MAE           & 6.89$\pm$0.186        & 13.34$\pm$0.293       & 19.24$\pm$0.345       & 24.55$\pm$0.186 \\
& & RMSE        & 10.22$\pm$0.305       & 19.20$\pm$0.402       & 27.04$\pm$0.439       & 33.83$\pm$0.523 \\

\cline { 3 - 7 } 
\multirow{2}{*}{3} & \multirow{2}{*}{1}
& MAE           & 6.82$\pm$0.033        & 13.41$\pm$0.195       & 19.47$\pm$0.446       & 24.62$\pm$0.484 \\
& & RMSE        & 10.21$\pm$0.106       & 19.27$\pm$0.266       & 27.29$\pm$0.555       & 33.90$\pm$0.512 \\

\cline { 3 - 7 } 
\multirow{2}{*}{3} & \multirow{2}{*}{2}
& MAE       & 6.71$\pm$0.119        & 13.24$\pm$0.221           & 19.30$\pm$0.181       & \textbf{24.14$\pm$0.214} \\
& & RMSE    & 10.05$\pm$0.189       & 19.07$\pm$0.291           & 27.05$\pm$0.397       & \textbf{33.36$\pm$0.459} \\

\cline { 3 - 7 } 
\multirow{2}{*}{3} & \multirow{2}{*}{3}
& MAE       & 7.71$\pm$0.219      & 13.58$\pm$0.376         & 19.63$\pm$0.580         & 24.74$\pm$0.567 \\
& & RMSE    & 10.70$\pm$0.273     & 19.60$\pm$0.449         & 27.71$\pm$0.831         & 34.18$\pm$0.721\\
\bottomrule
\end{tabular}}
\end{table*}

We implement our method with PyTorch \cite{paszke2019pytorch}, and construct GNNs with PyTorch Geometric library \cite{fey2019fast}. The code is released on GitHub\footnote{https://github.com/Friger/HighAir}. A server with one CPU (Intel Xeon Platinum 8163), 32GB RAM, and one GPU (NVIDIA Tesla V100) accomplishes all computing tasks, including training, validation, and test. To statistically measure the significance of performance differences, pairwise t-tests at 95\% significance level are conducted between HighAir and the compared methods in the following experiments ($*$ denotes the difference is statistically significant).

\subsection{Hyper-Parameter Evaluation}
We use distance weight $\lambda$ to control the number of edges. Parameter $\lambda$ is shared among all the graph constructions. Based on the definition in Section III-C, the smaller the value of $\lambda$ is, the less the number of edges is, but the adaptive method ensures that all nodes have at least one edge connected to other nodes.

\begin{figure}
    \centering
    \includegraphics{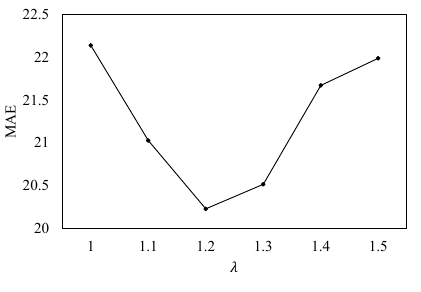}
    \caption{The evaluation results of distance weight $\lambda$.}
    \label{fig_9}
\end{figure}

We evaluate the effects of $\lambda$ with the average MAE of next couple of hours within forecasting horizon on validation data. Fig. \ref{fig_9} shows the evaluation results of different $\lambda$ values (step $=0.1$). We find that HighAir performs optimally when $\lambda$ equals to 1.2. As $\lambda$ increases, MAE decreases first and then increases. When $\lambda$ is too small, cities/stations cannot get sufficient information about adjacent cities/stations, while when $\lambda$ is too large, the local information about cities/stations would be over-diluted during the message passing process.

Within the testing range of $\lambda$ (i.e., from 1.0 to 1.5), although the MAEs do not have a significant variation, $\lambda$ has a great influence on the number of edges. For example, we conducted a test on Hangzhou city, and found that when $\lambda$ increases from 1.0 to 1.5, the number of edges almost doubles (increasing from 32 to 60).

We use $L_{\text {city}}$ and $L_{\text {station}}$ to denote the number of message passing layers at city-level and at station-level, respectively. To fully investigate the effects of $L_{\text {city}}$ and $L_{\text {station}}$, we conduct the hyper-parameter evaluation about these two hyperparameters, setting both $L_{\text {city}}$ and $L_{\text {station}}$ among \{1, 2, 3\}. Table \ref{tab:table_4} shows the evaluation results of different $L_{\text {city}}$ and $L_{\text {station}}$ values. We find that setting $L_{\text {city}}= $ 1 and $L_{\text {station}}= $ 1 achieves the best forecasting results for short and mid-term predictions (1h, 3h, 6h), and setting $L_{\text {city}}= $ 3 and $L_{\text {station}}= $ 2 performs best for long-term prediction (12h). One of the possible reasons for this phenomenon is that air pollutants take time to disperse from far away. Stacking more layers of message passing can expand the perceptual field for stations to receive the information from distant cities/stations, which facilitate long-term prediction. Considering the computation efficiency and time consumption, we set $L_{\text {city}}= $ 1 and $L_{\text {station}}= $ 1 in the experiments.

\begin{table*}[]
\centering
\caption{The results of factor evaluation (mean±std)}
\label{tab:table_5}
\resizebox{0.65 \textwidth}{!}{
\begin{tabular}{cccccc}
\toprule
Method & Metric     & 1h        & 3h        & 6h        & 12h \\
\midrule

\multirow{2}{*}{ HighAir } 
& MAE 
&\textbf{6.65}$\pm$\textbf{0.135}      &\textbf{13.15}$\pm$\textbf{0.382}         &\textbf{19.18}$\pm$\textbf{0.568}     &\textbf{24.16}$\pm$\textbf{0.583} \\
& RMSE 
&\textbf{9.97}$\pm$\textbf{0.193}      &\textbf{19.01}$\pm$\textbf{0.516}         &\textbf{26.97}$\pm$\textbf{0.821}     &\textbf{33.41}$\pm$\textbf{0.759} \\

\cline { 2 - 6 } 
\multirow{2}{*}{w/o weather} 
& MAE       &6.93$\pm$0.130$^*$      &13.76$\pm$0.232$^*$         &19.98$\pm$0.442$^*$         &25.34$\pm$0.681 \\
& RMSE      &10.26$\pm$0.206        &19.78$\pm$0.342$^*$         &28.10$\pm$0.577$^*$         &34.54$\pm$0.718$^*$ \\
\cline { 2 - 6 } 

\multirow{2}{*}{w/o poi} 
& MAE   &6.81$\pm$0.094          &13.27$\pm$0.222       &19.27$\pm$0.367      &24.52$\pm$0.454 \\
& RMSE  &10.21$\pm$0.237        &19.21$\pm$0.368      &27.18$\pm$0.634       &33.74$\pm$0.806 \\
\bottomrule
\end{tabular}}
\end{table*}

\begin{table*}[]
\centering
\caption{The results of model component evaluation (mean±std)}
\label{tab:table_6}
\resizebox{.65\textwidth}{!}{
\begin{tabular}{cccccc}
\toprule
Method & Metric & 1h        & 3h        & 6h       & 12h \\
\midrule

\multirow{2}{*}{ HighAir } 
& MAE 
&\textbf{6.65}$\pm$\textbf{0.135}      &\textbf{13.15}$\pm$\textbf{0.382}        &19.18$\pm$0.568    &24.16$\pm$0.583 \\
& RMSE 
&\textbf{9.97}$\pm$\textbf{0.193}      &\textbf{19.01}$\pm$\textbf{0.516}        &26.97$\pm$0.821    &33.41$\pm$0.759 \\

\cline { 2 - 6 } 
\multirow{2}{*}{w/o hierarchy}
& MAE &7.59$\pm$0.094$^*$       &14.35$\pm$0.225$^*$        &20.87$\pm$0.436$^*$       &26.32$\pm$0.485$^*$ \\
& RMSE &11.38$\pm$0.210$^*$     &20.23$\pm$0.338$^*$        &28.73$\pm$0.726$^*$       &35.06$\pm$0.670$^*$ \\

\cline { 2 - 6 } 
\multirow{2}{*}{w/o city LSTM}
& MAE       &6.83$\pm$0.160         &13.40$\pm$0.399      &19.30$\pm$0.453     &24.49$\pm$0.564 \\
& RMSE      &10.18$\pm$0.185      &19.26$\pm$0.482      &27.12$\pm$0.463      &33.59$\pm$0.668 \\

\cline { 2 - 6 } 
\multirow{2}{*}{w/o dynamic}
& MAE       &6.97$\pm$0.143$^*$        &13.44$\pm$0.188      &19.64$\pm$0.276      &24.97$\pm$0.374$^*$ \\
& RMSE      &10.36$\pm$0.261$^*$        &19.45$\pm$0.252     &27.61$\pm$0.377      &34.54$\pm$0.507$^*$ \\

\cline { 2 - 6 } 
\multirow{2}{*}{with GIN}
& MAE       &7.32$\pm$0.144$^*$       &13.58$\pm$0.301     &\textbf{19.10}$\pm$\textbf{0.566}     &\textbf{23.82}$\pm$\textbf{0.864} \\
& RMSE      &11.11$\pm$0.271$^*$     &19.48$\pm$0.381     &\textbf{26.81}$\pm$\textbf{0.836}       &\textbf{32.87}$\pm$\textbf{1.117} \\

\cline { 2 - 6 } 
\multirow{2}{*}{with GAT}
& MAE   &6.91$\pm$0.083$^*$      &13.61$\pm$0.302      &19.71$\pm$0.618      &24.70$\pm$0.645 \\
& RMSE  &10.24$\pm$0.140$^*$     &19.54$\pm$0.427      &22.77$\pm$0.949      &34.07$\pm$1.023 \\

\bottomrule
\end{tabular}}

\end{table*}

\subsection{Multi-Source Factor Evaluation}
To explore the effectiveness of multi-source factors, we compare HighAir with two variants, each one of which removes one kind of factors. Specifically, HighAir w/o weather removes weather data $\boldsymbol{weather}$ and HighAir w/o poi removes POI data $\boldsymbol{poi}$.

The performances of HighAir and variants are given in Table \ref{tab:table_5}, and we can draw the following conclusions:

(1) HighAir outperforms the other two variants, indicating that all factors can improve the performance of air quality forecasting.

(2) The ranking of the effectiveness of factors is weather data $>$ POI data. The result shows that air quality is mostly impacted by weather conditions.

\subsection{Model Component Evaluation}
To verify the effectiveness of different components, we compare HighAir with the following variants: (1) HighAir w/o hierarchy, which removes the city-level graph and the corresponding interactions, i.e., the city representation is removed from the global attributes of the station-level graphs; (2) HighAir w/o city LSTM, which removes the city LSTMs used in UD, and takes the mean AQI value of all stations in a city as the node attribute on the city-level graph; (3) HighAir w/o dynamic, which removes direction similarity from edge weights and introduces wind direction vector $\boldsymbol{wind}$ into the global attributes of the station-level graphs. (4) HighAir with other advanced GNNs, which replaces the adopted message passing mechanism with GIN \cite{xu2018powerful} and GAT \cite{velivckovicgraph}.

The performances of HighAir and variants are given in Table \ref{tab:table_6}, and the following tendencies can be discovered:

(1) HighAir outperforms HighAir w/o hierarchy on all metrics. The result indicates that the air quality of adjacent cities is beneficial, which can be used to model the diffusion processes of air pollutants from adjacent cities.

(2) HighAir outperforms HighAir w/o city LSTM on all metrics. The result indicates that the city LSTMs used in UD are effective, which can make cities perceive the historical air quality information of adjacent cities by the message passing process.

(3) HighAir outperforms HighAir w/o dynamic on all metrics. The result indicates that compared with taking wind direction as a feature contained in the global attributes, using it
to dynamically adjust the weights of edges is a more effective strategy, which can explicitly model the effect patterns of wind direction on air pollutant diffusion.

(4) HighAir outperforms HighAir with GAT on all metrics, but slightly inferior to HighAir with GIN for mid-term and long-term predictions (6h, 12h). The result indicates that our adopted message passing mechanism is effective, especially for the short-term predictions (1h, 3h). In addition, other possible implements can replace our adopted message passing mechanism for specific needs, e.g. using GIN for mid-term and long-term predictions (6h, 12h).

\subsection{Comparison with Other Forecasting Methods}
To further verify the effectiveness of our method, we compare HighAir with the following forecasting methods:

(1) HighAir with SAGE (HWS): HWS replaces the mean operation of upper delivery with a graph embedding method called SAGE \cite{li2019semi}, which considers the graph structure and learns the weights of different stations automatically.

(2) Historical Average (HA): HA takes the average value of historical AQIs as the forecasting result. Considering daily and weekly periodicity, we set the historical period to 168 (24×7).

(3) ARIMA: ARIMA consists of three time-series analysis parts: the auto-regressive part, the moving average part, and the integrate  part,  which  forecasts  the  air  quality  based  on  the historical AQIs of each station.

\begin{table*}[]
\centering
\caption{The results of different forecasting methods (mean±std)}
\label{tab:table_7}
\resizebox{0.65\textwidth}{!}{ 
\begin{tabular}{cccccc}
\toprule
Method & Metric & 1h & 3h & 6h & 12h \\
\hline

\multirow{2}{*}{ HighAir } 
& MAE & \textbf{6.65}$\pm$\textbf{0.135} & \textbf{13.15}$\pm$\textbf{0.382}& \textbf{19.18}$\pm$\textbf{0.568} & \textbf{24.16}$\pm$\textbf{0.583} \\
& RMSE & \textbf{9.97}$\pm$\textbf{0.193} & \textbf{19.01}$\pm$\textbf{0.516} & \textbf{26.97}$\pm$\textbf{0.821} & \textbf{33.41}$\pm$\textbf{0.759} \\

\cline { 2 - 6 } 
\multirow{2}{*}{HWS}
& MAE & 8.64$\pm$0.027$^*$ & 16.18$\pm$0.053$^*$ & 21.80$\pm$0.074$^*$  & 24.16$\pm$0.637 \\
& RMSE & 11.96$\pm$0.077$^*$ & 22.57$\pm$0.172$^*$ & 29.65$\pm$0.793$^*$  & 33.93$\pm$1.828 \\

\cline { 2 - 6 } 
\multirow{2}{*}{HA}
& MAE & 35.19$\pm$0.000$^*$ & 35.19$\pm$0.000$^*$ & 35.19$\pm$0.000$^*$ & 35.19$\pm$0.000$^*$ \\
& RMSE & 46.61$\pm$0.000$^*$ & 46.61$\pm$0.000$^*$ & 46.61$\pm$0.000$^*$ & 46.61$\pm$0.000$^*$ \\
\cline { 2 - 6 } 
\multirow{2}{*}{ARIMA}
& MAE & 9.55$\pm$0.000$^*$ & 19.72$\pm$0.000$^*$ & 26.13$\pm$0.000$^*$ & 31.97$\pm$0.000$^*$ \\
& RMSE & 13.13$\pm$0.000$^*$ & 27.75$\pm$0.000$^*$ & 35.02$\pm$0.000$^*$ & 42.25$\pm$0.000$^*$ \\
\cline { 2 - 6 } 
\multirow{2}{*}{GC-DCRNN}
& MAE & 8.55$\pm$0.028$^*$ & 16.80$\pm$0.021$^*$ & 22.77$\pm$0.102$^*$ & 27.23$\pm$0.731$^*$ \\
& RMSE & 11.51$\pm$0.442$^*$ & 22.86$\pm$0.339$^*$ & 30.77$\pm$0.712$^*$ & 36.52$\pm$1.591$^*$ \\
\cline { 2 - 6 }
\multirow{2}{*}{GC-LSTM}
& MAE & 9.05$\pm$0.013$^*$ & 17.05$\pm$0.075$^*$ & 22.76$\pm$0.155$^*$ & 26.97$\pm$1.370$^*$ \\
& RMSE & 12.08$\pm$0.104$^*$ & 22.94$\pm$0.312$^*$ & 30.54$\pm$0.614$^*$ & 36.09$\pm$0.528$^*$ \\
\cline { 2 - 6 }
\multirow{2}{*}{AIRMGCN}
& MAE & 8.98$\pm$0.126$^*$ & 16.29$\pm$0.591$^*$ & 22.09$\pm$0.898$^*$ & 26.49$\pm$0.787$^*$ \\
& RMSE & 12.29$\pm$0.171$^*$  & 23.08$\pm$0.673$^*$ & 30.65$\pm$0.439$^*$ & 36.12$\pm$0.648$^*$ \\
\cline { 2 - 6 } 
\multirow{2}{*}{HRN}
& MAE & 9.27$\pm$0.007$^*$  & 17.98$\pm$0.065$^*$ & 24.41$\pm$0.204$^*$ & 29.05$\pm$1.402$^*$ \\
& RMSE & 13.09$\pm$0.240$^*$ & 26.03$\pm$0.558$^*$ & 33.23$\pm$1.823$^*$ & 37.90$\pm$0.963$^*$ \\
\cline { 2 - 6 } 
\multirow{2}{*}{ST-UNet}
& MAE & 8.75$\pm$0.049$^*$ & 16.47$\pm$0.057$^*$  & 22.53$\pm$0.119$^*$ & 25.58$\pm$0.511$^*$ \\
& RMSE & 12.31$\pm$0.364$^*$ & 22.51$\pm$0.327$^*$ & 30.46$\pm$1.141$^*$ & 35.65$\pm$1.027$^*$ \\
\cline { 2 - 6 } 
\multirow{2}{*}{\wu{HGCN}}
& MAE & 7.35$\pm$0.249$^*$ & 15.64$\pm$0.217$^*$  & 20.53$\pm$0.819$^*$ & 25.38$\pm$0.391$^*$ \\
& RMSE & 11.31$\pm$0.364$^*$ & 21.21$\pm$0.226$^*$ & 27.46$\pm$0.748$^*$ & 34.95$\pm$1.031$^*$ \\
\bottomrule
\end{tabular}}
\end{table*}

(4) GC-DCRNN: GC-DCRNN \cite{lin2018exploiting} focuses on city wide air quality forecasting, and constructs the graph based on the POIs and the historical AQIs of stations. It combines recurrent neural networks with diffusion convolution to forecast the air quality. To keep the fairness of comparison, we constructed the graph based on the stations of all cities, and introduce weather data into the input of the decoder.

(5) GC-LSTM: GC-LSTM \cite{qi2019hybrid} constructs a flat graph based on the stations of all cities, and utilizes graph convolution operation and LSTM to model spatial and temporal dependencies, respectively. To keep the fairness of comparison, we introduce POI data into node attributes and weather data into the input of the decoder.

(6) AIRMGCN: AIRMGCN follows STMGCN \cite{geng2019spatiotemporal}, which is a spatial-temporal forecasting model based on multiple graphs. For station-level air quality forecasting, we constructed two graphs based on the spatial distances between stations and the cities they belong to, respectively. To keep the fairness of comparison, we improve the model to enable multi-step forecasting.

(7) HRN: HRN \cite{mrowca2018flexible} is a hierarchical model utilizes FNNs to model the inter-level and intra-level interactions. HRN implements all the interactions at the last time slot in the historical time window. To keep the fairness of comparison, we improve the model to enable multi-step forecasting, and introduce POI data and weather data into node attributes.

(8) ST-UNet: ST-UNet \cite{yu2019st} is a spatial-temporal forecasting method that leverages pooling operation to coarsen a graph in spatial domain and adopts dilated RNN to capture temporal dependencies. To keep the fairness of comparison, we introduce POI data into node attributes, weather data into global attributes, and weather data into the input of the decoder.

\wu{(9) HGCN: HGCN \cite{guo2021hierarchical} is a hierarchical spatial-temporal forecasting model that constructs the station-level graph structure based on Euclidean distances and conducts spectral clustering on the Laplacian matrix of the station-level graph to obtain the hierarchy. It applies GCN and Attention to capture spatial and temporal dependencies, respectively. To keep the fairness of comparison, we introduce POI data and weather data as covariates for air quality forecasting.}

The performances of HighAir and other forecasting methods are given in Table \ref{tab:table_7}, and the following tendencies can be discovered:

(1) HighAir outperforms HWS in short-term forecasting, but HWS performs better in long-term forecasting. It indicates that the topologies of the station-level graphs are helpful for long-term forecasting but have little effects on short-term forecasting, as they contain information that is more conducive to long-term forecasting, e.g., the distribution of stations and pollutant diffusion paths. For short-term forecasting, a non-parametric method performs better with limited training data.

(2) HighAir outperforms HA and ARIMA in both short-term and long-term forecasting. It indicates that a method only based on time-series analysis models cannot effectively forecast air quality, as air quality is affected by multi-source complex factors.

(3) HighAir outperforms GC-DCRNN, GC-LSTM, and AIRMGCN. It indicates that a hierarchical structure can model spatial dependencies more effectively than a flat structure. The reason might be that the messages between the stations located in different cities are hard to pass on a flat graph, as the distances between these stations are too large as compared with the distances between adjacent stations.

(4) HighAir outperforms HRN. It indicates that the encoder-decoder architecture and GNN are effective in capturing dependencies in temporal and spatial dimensions, as the encoder-decoder architecture and GNN have advantages in the modeling of sequences and non-Euclidean distributed entities, respectively.

(5) HighAir outperforms ST-UNet. It indicates that a stable HGNN based method performs better in air quality forecasting. It might be because after coarsening operation, the nodes in the coarsened graph no longer correspond to entities, which cannot support city-level message passing.

\wu{(6) HighAir outperforms HGCN. It indicates that considering dynamic factors to adjust the hierarchy may help to capture time-varying dependencies between cities.}

\subsection{Case Study}
In this case study, we try to demonstrate that our method gains more advantage over the existing methods when the wind factor between cities is more significant. The task is to forecast the AQIs of Shanghai city from 13:00 December 27 to 0:00 December 28, 2018. Table \ref{tab:table_8} shows the true AQIs (the mean AQI of all stations in a city) of Shanghai city and its connected cities on the city-level graph, i.e., Jiaxing, Suzhou, and Nantong, from 13:00 December 26 to 12:00 December 27, 2018, every four hours. In addition, $ws$ denotes the wind direction similarity (Section III-C) between each city and Shanghai city.

\begin{table}[]
\centering
\caption{The air quality of Shanghai city and adjacent cities from December 26 to 27, 2018}
\label{tab:table_8}
\resizebox{0.45\textwidth}{!}{ 
\begin{tabular}{@{}cccccccc@{}}
\toprule
\multirow{3}{*}{City} & \multirow{3}{*}{Data} & \multicolumn{6}{c}{Historical Time Window} \\ \cmidrule(l){3-8} 
                      & \multicolumn{1}{r}{}                      & 16:00 & 20:00 & 0:00 & 4:00 & 8:00 & 12:00 \\ 
                        \midrule
Shanghai              & AQI                                       & 45.5  & 61.2  & 75.3 & 74.9 & 75.2 & 101.5 \\ 
                        \cmidrule(l){2-8} 
\multirow{2}{*}{Jiaxing }              & AQI                                       & 95.5  & 67    & 86.5 & 93   & 68   & 115.5 \\
                      & ws                                        & -0.2  & -0.2  & -0.98     & -0.98     &-0.98      &-0.2       \\ 
                      \cmidrule(l){2-8} 
\multirow{2}{*}{Nantong}               & AQI                                      &53.8	 &151.6	&129.2	&154.0	&180.2	&213.4       \\
                      & ws                                       &-0.12  &-0.12	&0.99	&0.99	&-0.99	&0.12       \\ 
                      \cmidrule(l){2-8} 
\multirow{2}{*}{Suzhou}                & AQI                                      &87.1	&146.1   &153.3    &158.1	&103.4	&148.9       \\
                      & ws                                       &0.65  &-0.76   &-0.76	&0.65	&0.65	&0.76 \\ 
                      \bottomrule
\end{tabular}
}
\end{table}

In this case, the air qualities of Nantong city and Suzhou city are worse than Shanghai city. At most of the time slots, wind direction similarity between Shanghai city and Nantong city, as well as Suzhou city, is greater than 0, which means that the air pollutants from these two cities would be brought to Shanghai city by wind. This is a typical air quality deterioration case, which is caused by the air pollutants diffused from adjacent cites.

We compare the performances of HighAir with other forecasting methods in Table \ref{tab:table_9} (the average MAE of the stations located in Shanghai city are reported). HighAir w/o N\&S is a variant of HighAir, which sets ws between Nantong city \& Shanghai city and Suzhou city \& Shanghai city as 0 (i.e., the wind factor is ignored). We can find that the short-term forecasting errors become larger than usual, as the dependencies in spatial and temporal dimensions are complicated when air quality suddenly deteriorates. Compared with other methods, HighAir achieves high accuracy in both short-term and long-term forecasting tasks. This justifies that our hierarchical structure design and the dynamic adjustment of edge weights can well model the diffusion processes of air pollutants from adjacent cities. In particular, HighAir outperforms HighAir w/o N\&S. It indicates that our method could capture the wind factor between cities, which has a great effect in this case.

\section{Conclusions and Future Work}

\begin{table}[]
\centering
\caption{The names of different methods in the case}
\resizebox{.4\textwidth}{!}{
\label{tab:table_9}
\begin{tabular}{ccccc}
\toprule
Methods & 1h & 3h & 6h & 12h \\
\midrule
HighAir & \textbf{15.33} & \textbf{26.12} & \textbf{29.64} & \textbf{32.12} \\
HighAir w/o N\&S & 18.87 & 38.12 & 40.02 & 43.09 \\
GC-DCRNN & 20.24 & 35.85 & 38.83 & 41.42 \\
GC-LSTM & 21.30 & 41.18 & 41.37 & 44.82 \\
AIRMGCN & 16.96 & 30.54 & 36.32 & 40.02 \\
HRN & 19.22 & 48.65 & 50.19 & 48.19 \\
ST-UNet & 20.30 & 34.71 & 37.03 & 36.51 \\
\wu{HGCN} & 18.93 & 27.07 & 32.24 & 35.16 \\
\bottomrule
\end{tabular}}
\end{table}

We propose HighAir, a hierarchical graph neural network-based air quality forecasting method. HighAir adopts an encoder-decoder architecture and considers complex air quality influencing factors, e.g., weather and land usage. Specifically, we construct a city-level graph and station-level graphs from a hierarchical perspective, which can consider city-level and station-level patterns, respectively. We design two strategies, i.e., upper delivery and lower updating, to implement the inter-level interactions, and adopt message passing mechanism to implement the intra-level interactions. To evaluate the proposed method, experiments have been conducted on a real-world dataset, and the results demonstrate the effectiveness of method components and the superior performance of HighAir.

In the future, we will extend our method in the following aspects. First, we have discovered that the same graph embedding method has different effects on short-term and long-term forecasting tasks (Section IV-E), so we will study strategies (e.g., attention mechanism and meta learning) to automatically learn the importance of topological information for different forecasting horizons. Second, HighAir utilizes GNN to capture the spatial dependencies, but the actual diffusion processes of air pollutants is more complex than the message passing process on graphs. So we will try to leverage semi-supervised learning to utilize the information of the regions without monitoring stations. In addition, we will validate HighAir on datasets with larger graph sizes. From a theoretical analysis, deepening graph neural networks to capture the spatial dependencies between distant entities, e.g., two stations in different cities, would cause the problem of over-smoothing \cite{liu2020towards}. Utilizing graph pooling method to capture multi-scale representations has become a common practice to model the large graphs \cite{ying2018hierarchical}, \cite{lee2019self}. Similarly, HighAir captures spatial dependencies from station-level and city-level perspectives, which would perform better on datasets with larger graph sizes.

\ifCLASSOPTIONcaptionsoff
  \newpage
\fi
\fontsize{14}{14}{
\bibliography{IEEEfull}
\bibliographystyle{IEEEtran}

\begin{IEEEbiography}
[{\includegraphics[width=1in,height=1.25in,clip,keepaspectratio]{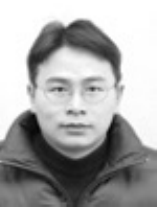}}]
{Ling Chen }
received the B.S. and Ph.D. degrees in computer science from Zhejiang University, China, in 1999 and 2004, respectively. He is currently a Professor with the College of Computer Science and Technology, Zhejiang University, China. His research interests include ubiquitous computing and data mining.
\end{IEEEbiography}

\begin{IEEEbiography}
[{\includegraphics[width=1in,height=1.25in,clip,keepaspectratio]{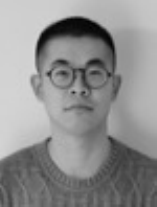}}]
{Jiahui Xu}
received the B.Eng. degree in Communication Engineering from Wuhan University of Technology, China, in 2018. He is currently a M.S. candidate with the College of Computer Science and Technology, Zhejiang University, China. His research interests include urban computing and time series modeling.
\end{IEEEbiography}

\begin{IEEEbiography}
[{\includegraphics[width=1in,height=1.25in,clip,keepaspectratio]{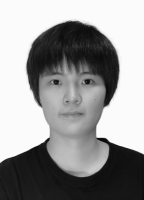}}]
{Binqing Wu}
received the B.Eng. degree in computer science from Southwest Jiangtong University, China, in 2020, and the B.Eng. degree in computer science from University of Leeds, UK, in 2020. She is currently a Ph.D. student with the College of Computer Science and Technology, Zhejiang University, China. Her research interests include time series forecasting and data mining.
\end{IEEEbiography}

\begin{IEEEbiography}
[{\includegraphics[width=1in,height=1.25in,clip,keepaspectratio]{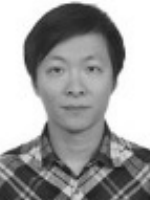}}]
{Mingqi Lv}
received the Ph.D. degree in computer science from Zhejiang University, Hangzhou, China, in 2012. He is currently an Associate Professor with the College of Computer Science and Technology, Zhejiang University of Technology, China. His research interests include spatiotemporal data mining and ubiquitous computing.
\end{IEEEbiography}

\begin{IEEEbiography}
[{\includegraphics[width=1in,height=1.25in,clip,keepaspectratio]{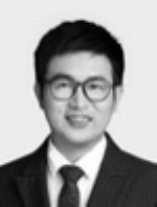}}]
{Chaoqun Zhan}
is currently a director of Alibaba Database Business. His research interests include distributed computing and cloud computing. He developed the large-scale online data analysis products AnalyticDB and Data Lake Analytics from scratch.
\end{IEEEbiography}

\begin{IEEEbiography}
[{\includegraphics[width=1in,height=1.25in,clip,keepaspectratio]{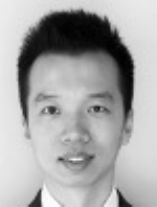}}]
{Sanjian Chen}
received the Ph.D. degree in computer and information science from the University of Pennsylvania in 2016. He is currently a Senior Algorithm Expert at the Alibaba Group. His research interests include large-scale machine learning algorithms.
\end{IEEEbiography}

\vspace{-440px}
\begin{IEEEbiography}
[{\includegraphics[width=1in,height=1.25in,clip,keepaspectratio]{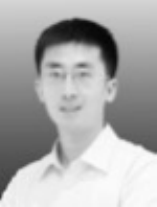}}]
{Jian Chang} received the Ph.D. degree in computer and information science from the University of Pennsylvania in 2013. He is currently a Senior Algorithm Expert at the Alibaba Group. His research interests include cutting-edge applications of AI at the intersection of high-performance databases and the IoT.
\end{IEEEbiography}
}

\end{document}